\documentclass[lettersize,journal]{IEEEtran}
\usepackage{amsmath,amsfonts}
\usepackage{algorithmic}
\usepackage{array}
\usepackage[caption=false,font=normalsize,labelfont=sf,textfont=sf]{subfig}
\usepackage{textcomp}
\usepackage{stfloats}
\usepackage{url}
\usepackage{verbatim}
\usepackage{graphicx}

\usepackage[backref]{hyperref}
\usepackage{amsmath,bm}
\usepackage{array}
\usepackage{comment}
\usepackage{multirow}
\usepackage{multicol} 
\usepackage{booktabs}
\usepackage{color}
\usepackage{pifont} 

\def\BibTeX{{\rm B\kern-.05em{\sc i\kern-.025em b}\kern-.08em
    T\kern-.1667em\lower.7ex\hbox{E}\kern-.125emX}}
\usepackage{balance}
\begin{document}
\title{Progressive Channel-Shrinking Network}
\author{Jianhong Pan, \and
Siyuan Yang, \and
Lin Geng Foo, \and
Qiuhong Ke, \and
Hossein Rahmani, \and
Zhipeng Fan, \and
Jun Liu
\thanks{Jianhong Pan, Lin Geng Foo and Jun Liu are with Singapore University of Technology and Design, Singapore.
Siyuan Yang is with Nanyang Technological University, Singapore.
Qiuhong Ke is with Monash University, Australia.
Zhipeng Fan is with New York University, USA.
Hossein Rahmani is with Lancaster University, UK.
(Corresponding author: Jun Liu)
}
}

\markboth{Journal of \LaTeX\ Class Files,~Vol.~18, No.~9, September~2020}%
{How to Use the IEEEtran \LaTeX \ Templates}

\maketitle

\begin{abstract}
Currently, salience-based channel pruning makes continuous breakthroughs in network compression. In the realization, the salience mechanism is used as a metric of channel salience to guide pruning. Therefore, salience-based channel pruning can dynamically adjust the channel width at run-time, which provides a flexible pruning scheme. However, there are two problems emerging: a gating function is often needed to truncate the specific salience entries to zero, which destabilizes the forward propagation; dynamic architecture brings more cost for indexing in inference which bottlenecks the inference speed. In this paper, we propose a Progressive Channel-Shrinking (PCS) method to compress the selected salience entries at run-time instead of roughly approximating them to zero. We also propose a Running Shrinking Policy to provide a testing-static pruning scheme that can reduce the memory access cost for filter indexing. We evaluate our method on ImageNet and CIFAR10 datasets over two prevalent networks: ResNet and VGG, and demonstrate that our PCS outperforms all baselines and achieves state-of-the-art in terms of compression-performance tradeoff. Moreover, we observe a significant and practical acceleration of inference. The code will be released upon acceptance.
\end{abstract}


\section{Introduction}
\IEEEPARstart{T}{he} The performance of convolutional neural networks (CNNs) has been significantly improved alongside the milestone architectures being proposed, including AlexNet~\cite{krizhevsky2012imagenet}, VGG~\cite{simonyan2014very}, GoogLeNet~\cite{szegedy2015going}, ResNet~\cite{he2016deep}, DenseNet~\cite{huang2017densely}, ResNeXt~\cite{xie2017aggregated}, and SE-Net~\cite{hu2018squeeze}. 
However, it comes with a price that CNNs become massive and thus inefficient, in which convolution operations contribute the major computational complexity (Multiply-Add operations denoted as MAdds). To reduce MAdds, the group convolution and its variants~\cite{huang2018condensenet, sun2018igcv3, xie2018interleaved, xie2017aggregated, zhang2017interleaved,wang2021spgnet} split channels into several parts as independent convolution branches and prune the connection among them. Moreover, depth-wise separable convolution~\cite{chollet2017xception,howard2017efficient,sandler2018inverted}, an extreme version of group convolution, prunes all channel-wise connections. Besides, shuffle operation~\cite{ma2018shufflenet, zhang2018shufflenet,liu2017learning,wen2016learning} is adopted as a low-cost operation to link channels. 
However, these pruning schemes are indiscriminate where both important and unimportant connections can be pruned together.

To achieve efficient pruning, we need clues to decide which channels should be pruned. 
Some works~\cite{li2016pruning,wen2016learning,liu2017learning,he2017channel,ye2018rethinking,liebenwein2019provable} evaluate channel salience measures such as the magnitude of the weights or activation functions after training and prune the relatively unimportant channels. To mitigate the accuracy drop from pruning, the network is retrained. These methods are called static pruning. 

Unlike static pruning approaches, the dynamic methods~\cite{chen2018gaternet,hua2018channel, lin2017runtime,bejnordi2019batch,chen2019self,gao2018dynamic,herrmann2018end,guo2020model,xia2021fully,su2020dynamic,bochkovskiy2020yolov4,tang2021manifold,li2021dynamic2,zhao2020efficient,ople2021adjustable,zhang2020accurate}, prune channels and fine tune the network simultaneously. Salience-based pruning is a recent running-time method that allows the network to learn the importance of channels from the input and the whole network status. In the realization, the salience reweighs the feature maps and those that are assigned zero weights will be pruned as the deactivated channels. Hence, salience-based pruning can dynamically adjust the pruning scheme, where the channel width is scaled elastically. Gating networks~\cite{chen2018gaternet,hua2018channel, lin2017runtime} and attention~\cite{bejnordi2019batch,chen2019self,gao2018dynamic,herrmann2018end,guo2020model,xia2021fully,su2020dynamic,bochkovskiy2020yolov4,tang2021manifold,li2021dynamic} are often used to predict the importance of channels and decide which channels should be dropped. 

However, there are two issues in these dynamic channel-wise pruning methods: 
\textbf{First}, the pruning operation is abrupt, which harms the network performance: Because the salience vector rarely contains zero entries, to deactivate the channels' outputs (channels pruning), a gate or a step function is often used to quantize/truncate some salience entries to zero. Quantization abruptly deactivates the channels in service. With various inputs, the pruning schemes are different during training, hence the abrupt pruning occurs all the time, leading to unstable training. 
\textbf{Second}, the inference becomes inefficient because the pruning scheme is not fixed for different input samples. The unfixed scheme means random channel indexing in response to different inputs. Hence, massive indexing during testing leads to a higher memory-access cost (MAC) and bottlenecks the inference speed. 

In this work, we propose a novel Progressive Channel-Shrinking (PCS) method to address the above problems: 
\textbf{1)} 
We use a salience generator without truncation for continuity and differentiability, avoiding the backward propagation problem. Then, we generate zero entries by progressively shrinking the salience entries associated with the relative low-salience channels. \textbf{2)}~We propose a Running Shrinking Policy to avoid massive weight indexing and significantly promote inference speed. Running Shrinking Policy guarantees an identical pruning scheme for all samples. After training, we can directly remove the deactivated channels and the indexing operation is no longer needed during inference. Besides, the Running policy maintains a dynamic scheme in training and achieves good performance. \textbf{3)}~We embed the PCS module into popular deep CNN models, such as ResNet \cite{he2016deep} and VGG~\cite{simonyan2014very}, and evaluate its effectiveness on ImageNet and CIFAR10 datasets. 
The experimental results indicate our PCS outperforms existing channel pruning methods.

\section{Related Work}
\label{Related Work}

Various channel pruning methods, including salience-based \cite{bejnordi2019batch,chen2019self,herrmann2018end,guo2020model,xia2021fully,su2020dynamic,bochkovskiy2020yolov4,tang2021manifold,li2021dynamic} and gating-based \cite{chen2018gaternet,hua2018channel, lin2017runtime} methods have been proposed recently. Both the salience-based and gating-based channel pruning are dynamic pruning methods because they respectively generate channel-based salience and gating vectors using a module or sub-network to prune unimportant channels according to the inputs, i.e., different convolutional channels are activated in response to various inputs. 
We remark that channel pruning methods belong to a more general category of feature selection methods (that include Principal Component Analysis and Linear Discriminant Analysis) which aim to extract important features from the data. More precisely, channel pruning can be considered as \textit{a type of feature selection method} that is specifically developed for deep neural networks. 
where the main purpose is to identify and preserve the most important channels (while removing the less important ones) to improve computational complexity of the deep neural network.

\textbf{Gating-Based Channel Pruning.}
Gating network outputs discrete one or zero as a switch to activate or deactivate a channel
~\cite{chen2018gaternet,hua2018channel,lin2017runtime}. Hua \textit{et al.}~\cite{hua2018channel} utilised a gating network to manage the subset of channels. Lin \textit{et al.}~\cite{lin2017runtime} used RNN with a gating function to select important channels. 
However, the non-differentiability of the gating network often leads to some backward propagation problems~\cite{han2021dynamic}. 
To address the backward propagation problem, Wang \textit{et al.}~\cite{wang2018skipnet} proposed a hybrid reinforcement learning method; Veit \textit{et al.}~\cite{veit2018convolutional} and Herrmann \textit{et al.}~\cite{herrmann2019end} adopted Gumbel SoftMax~\cite{gumbel1954statistical,jang2016categorical} to probabilize the gate.

\textbf{Salience-Based Channel Pruning.} 
Salience mechanism was first introduced for visual perception~\cite{corbetta2002control,itti1998model,rensink2000dynamic} and then used to dynamically scale the values of feature maps~\cite{hu2018squeeze,park2018bam,ulutan2020vsgnet,wang2017residual,woo2018cbam,xu2016ask}. Salience Mechanism can also be used to predict the importance of channels in convolutional networks. For example, Hu \textit{et al.}~\cite{hu2018squeeze} proposed squeeze-and-excitation (SE) module to generate channel-wise salience vector to reweigh the feature maps. Several other approaches also use the attention mechanism to predict the salience of channels as a guidance for pruning policy~\cite{bejnordi2019batch,chen2019self,gao2018dynamic,herrmann2018end,guo2020model,xia2021fully,su2020dynamic,bochkovskiy2020yolov4,tang2021manifold,li2021dynamic}, i.e., the salience generator predicts salience of channels~\cite{hu2018squeeze,tang2021manifold} and generates a channel-based salience vector to 
reweigh channels for pruning. In practice, the salience vector is quantized to generate zero entries to
deactivate its corresponding unimportant channels. For example, when an entry $s_c$ at the $c^{\text{th}}$ channel of the salience vector $\boldsymbol{s}=(s_{1},s_{2},...,s_{C})$ becomes zero, the corresponding feature map at the $c^{\text{th}}$ channel will be reweighed to zero. However, the salience generator is a continuous function, from which the output salience vector hardly contains zero entries. Hence, a truncation is often used to zero some relatively small entries:
\begin{align}
  z
  = \left\{
            \begin{array}{cl}
                z ,  & z\geq\eta \\
                0 , & z<\eta
            \end{array}
            \right. ,
\end{align}
where $\eta$ is the threshold. 
However, truncation is abrupt and leads to the values \emph{plummeting} to the bottom once they are below the threshold. During training, when an entry is truncated, the corresponding channel will be deactivated immediately when it is still on service and contributes information to the next convolution layer as well as the network output. This results in an abrupt increase of the loss, making the training unstable and deteriorating the network performance.

A few other works~\cite{gao2018dynamic,su2020dynamic,tang2021manifold} utilised LASSO~\cite{tibshirani1996regression} to compress the attention entries in order to set some of the small-value entries to zero. However, LASSO suppresses all the channels simultaneously and indiscriminately and then directly filters out low-saliency entries. During this process, the salient channels will also be suppressed, affecting the performance of the model. Moreover, the filtering process often involves truncation of the low-salient entries, which might still be active though less dominant. The truncation further introduces discontinuity, leading to the fluctuation of convergence. To address these issues, we propose Progressive Shrinking to \emph{gradually suppress only the low-salience channels} and finally turn them off completely using hard sigmoid.

Furthermore, channel indexing is required for each input to index out the selected channels during inference. This results in high computational complexity (MAdds) as well as high memory access cost (MAC), which bottleneck the inference speed \cite{ma2018shufflenet}. In this work, we propose a Running Shrinking Policy, which allows the pruning to be static during testing while remaining dynamic during training. Compared to the traditional dynamic pruning, our method significantly improves the practical inference speed at no discernible performance drop.

\begin{figure*}[t]
  \center
  \includegraphics[height=4.6cm]{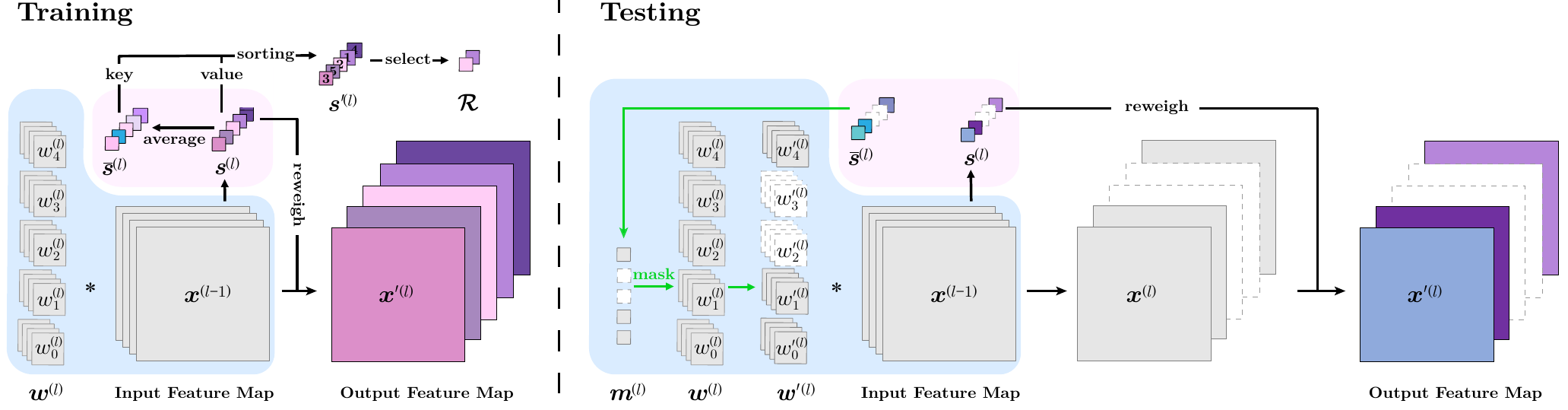}  
  \caption{The demonstration of training and testing of our PCS. $*$ denotes the convolution operation. $\boldsymbol{w}^{(l)}$, $\boldsymbol{w}^{\prime(l)}$, $\boldsymbol{x}^{(l-1)}$, and $\boldsymbol{x}^{(l)}$ denote the convolutional weights, the pruned weights, the input feature maps, and the output feature maps of the $l^{\text{th}}$ layer, respectively.
  $\boldsymbol{s}^{(l)}$ denotes the salience vector which reweighs the feature maps and $\overline{\boldsymbol{s}}^{(l)}$ denotes the moving average of $\boldsymbol{s}^{(l)}$. During the training, $\boldsymbol{s}^{(l)}$ is sorted based on the values of $\overline{\boldsymbol{s}}^{(l)}$ using the key-value mechanism. Then, selection is performed on the sorted $\boldsymbol{s}^{\prime (l)}$ for the construction of the Shrinking Loss $\mathcal{R}$. During inference, the Boolean mask $\boldsymbol{m}^{(l)}$ is generated from the running salience vector $\overline{\boldsymbol{s}}$ to mask and prune the weights. 
  Masking (\textcolor{green}{in green}) is only operated once after training to prune the low-salience weights permanently. 
  The masks, weights, and feature maps in white denote their values are zero. }
  \label{PCSConv}
\end{figure*}

\section{Progressive Channel-Shrinking}
\label{Progressive Channel-Shrinking}
Generally, the channel-wise pruning mechanism reweighs each channel of the output feature maps of a convolutional layer, as follows: 
\begin{align}
  \label{eq.attproduct}
  \boldsymbol{x}^{\prime}_{c}=s_{c}\boldsymbol{x}_{c}, 
\end{align}
where $\boldsymbol{x}_{c}\in\mathbb{R}^{H \times W}$ denotes  the output feature map of the $c^{\text{th}}$ channel.  $H$ and $W$ are the height and width of the output feature map, respectively.  $\boldsymbol{x}^{\prime}_{c}\in\mathbb{R}^{H \times W}$ denotes the reweighed feature map of the $c^{\text{th}}$ channel, and $s_{c}\in \mathbb{R}$ denotes the corresponding 
salience entry, which represents the importance of the  $c^{\text{th}}$ channel. 
Note that, once the 
salience entry becomes zero,
$\boldsymbol{x}'_{c}$ will become zero accordingly, which indicates that the $c^{\text{th}}$ channel is deactivated. 
Hence, pruning unimportant channels can be achieved by zeroing low-salience entries.

However, as mentioned in Sec. \ref{Related Work}, many salience-based channel-wise pruning models use truncation to generate zero-valued entries, which leads to unstable training and limited network performance (details in Sec.~\ref{sec: Ablation Study}). Moreover, the salience entries (e.g., $s_{c}$) depend on individual inputs, and thus the distributions of zero-value entries vary in response to different inputs during inference.
This leads to a massive channel indexing cost for different inputs, which significantly increases MAC \cite{ma2018shufflenet} and slows down the inference speed.

To address these issues, we introduce a Progressive Channel-Shrinking (PCS) method, which consists of a Progressive Shrinking strategy to progressively trim unimportant channels to zero along the training instead of roughly truncating them, and a Running Shrinking Policy to force the selected pruning channels stable for different inputs to avoid costly channel-indexing in inference.

The overall framework of the proposed method is shown in Fig. \ref{PCSConv}. More specially, the $l^{\text{th}}$ convolutional layer takes the output feature maps from the previous layer $\boldsymbol{x}^{(l-1)}$ as its input feature maps. 
$\boldsymbol{x}^{(l-1)}$ is used  to generate a salience vector $\boldsymbol{s}^{(l)}=(s^{(l)}_{1},s^{(l)}_{2},...,s^{(l)}_{C})$, which is progressively shrunk to generate zero-value entries  to reweigh the output feature maps $\boldsymbol{x}^{(l)}$ with $C$ channels for channel pruning (Sec.~\ref{Sec: Progressive Shrinking}). To construct an input agnostic pruning mask to reduce the MAC, we further introduce a running salience vector $\overline{\boldsymbol{s}}^{(l)}$ based on the history of $\boldsymbol{s}^{(l)}$. In return, the running salience vector $\overline{\boldsymbol{s}}^{(l)}$ further guides the updates of the salience vector $\boldsymbol{s}^{(l)}$ and also fine tunes the target model to adapt to the pruned weights (Sec.~\ref{Sec: Running Shrinking Policy}).   
Finally, $\overline{\boldsymbol{s}}^{(l)}$ is converted to a Boolean mask $\boldsymbol{m}^{(l)}$ to prune convolution weights $\boldsymbol{w}^{(l)}$ before the convolution operation to reduce the computation cost (Sec.~\ref{Sec: Accelerating Inference Process}). With the stable $\overline{\boldsymbol{s}}^{(l)}$ and $\boldsymbol{m}^{(l)}$ that are irrelevant to individual inputs, the channel-indexing operation can be done before deployment. Hence, the practical inference speed is significantly improved after training.

\subsection{Progressive Shrinking}
\label{Sec: Progressive Shrinking}

In this section, we propose a shrinking loss, which \textit{progressively} in every backward propagation, shrinks the $K$ lowest-salience entries of the salience vector at each layer to zero, instead of roughly truncating them. Unlike LASSO-based works~\cite{gao2018dynamic,su2020dynamic,tang2021manifold} which suppress all the channels simultaneously and indiscriminately, our proposed shrinking loss gradually suppresses only the low-salience channels leading to stable training and better network performance. 

To select the lowest-salience entries, we sort the
salience entries $\boldsymbol{s}$ in a monotonic increasing order to get $\boldsymbol{s}'$ (here the notion of layers is omitted to simplify representation). We then define the shrinking loss as follows:
\begin{align}
\label{eq: Shrinking Loss}
  \mathcal{R} = \sum_{i\leq K} s'_i
\end{align}
where ${s_i}'$  denotes the $i^{\text{th}}$ entry of  $\boldsymbol{s}'$. $K$ is a pre-defined free hyper-parameter, which determines the number of channels for pruning. 
$\boldsymbol{s}$ is obtained by feeding the input feature maps of a convolutional layer to a salience generator.
Hence the entries of $\boldsymbol{s}$ are  non-negative and the shrinking will stop if the selected entries become zero.
In summary, the overall hybrid learning objective is defined as:
\begin{equation}
    \begin{aligned}
  \label{overall objective}
  &\mathop{\arg\min}_{\theta,\pi}\mathcal{J}(F_{\theta},G_{\pi})= \\
  &
  \mathop{\arg\min}_{\theta,\pi}\mathbb{E}_{\boldsymbol{x}^{(0)}}\Big[\mathcal{L}(\boldsymbol{\hat{y}} (\boldsymbol{x}^{(0)},F_{\theta},G_{\pi}),\boldsymbol{y})+\lambda \mathcal{R}(\boldsymbol{x}^{(0)},F_{\theta},G_{\pi})\Big], 
\end{aligned}
\end{equation}
where $\boldsymbol{x}^{(0)}$ denotes the network input, i.e., the input feature maps of the first layer, $\theta$ and $\pi$ are respectively the parameters of the network $F_{\theta}$ and the 
salience generator $G_{\pi}$, $\mathcal{L}$ is the task loss of the network $F_{\theta}$ (e.g. cross-entropy loss to measure network classification performance), $\mathcal{R}$ is the combined per-layer shrinking loss in Eq.~\eqref{eq: Shrinking Loss}, $\boldsymbol{\hat{y}}$ is the estimate of the ground truth $\boldsymbol{y}$, and $\lambda$ is the shrinking rate at each optimizing step. To ensure that the salience entries corresponding to the low-salience channels can be shrunk to zero, we gradually increase $\lambda$ (see details in Sec. \ref{Sec: Experiments}). After training, a salience vector with $K$ zero-valued entries can be generated for each input sample. Hence, we can prune the corresponding $K$ channels to achieve lower MAdds.

According to Eq. \eqref{overall objective}, we can get the final shrinkage of the salience entry $s_i$ in the $l^{\text{th}}$ layer as: 
\begin{equation}
    \begin{aligned}
  \nabla_{s_i} \mathcal{J}(F_{\theta},G_{\pi})&=
  \mathbb{E}_{\boldsymbol{x}^{(0)}}\nabla_{s_i} (\mathcal{L}+\lambda \mathcal{R})\\
  &=\mathbb{E}_{\boldsymbol{x}^{(0)}}\nabla_{s_i} \mathcal{L}+\lambda\mathbb{E}_{\boldsymbol{x}^{(0)}}\nabla_{s_i}\sum_{j\leq K} s'_j\\
  &=\left \{
    \begin{array}{ll}
      \mathbb{E}_{\boldsymbol{x}^{(0)}}\nabla_{s_i} \mathcal{L}+\lambda\mathbb{E}_{\boldsymbol{x}^{(0)}}\frac{\partial s_i}{\partial s_i}, & s_i\in R,\\
      \mathbb{E}_{\boldsymbol{x}^{(0)}}\nabla_{s_i} \mathcal{L} ,     & s_i\notin R,
    \end{array}
  \right.\\
  &=\left \{
    \begin{array}{ll}
      \mathbb{E}_{\boldsymbol{x}^{(0)}}\nabla_{s_i} \mathcal{L}+\lambda, & s_i\in R,\\
      \mathbb{E}_{\boldsymbol{x}^{(0)}}\nabla_{s_i} \mathcal{L} ,     & s_i\notin R,
    \end{array}
  \right.
\label{eq:PCSGrad}
\end{aligned}
\end{equation}
where $R=\{s'_i|i\leq K\}$ contains all the selected lowest-salience entries.
Here, $\lambda$ is very small at the beginning and increases over training iterations. Hence, the task loss $\mathcal{L}$ dominates the optimization of the network in the early stage of training, where the top-K selection is dynamic and self-adaptive. As $\lambda$ increases, the shrinking loss starts to dominate the optimization where the network tends to select and shrink the lowest-salience entries, and the selection gradually becomes stable. For the $K$ lowest-salience entries ($s_i\in R$), the gradient is gradually increased with the increasing  $\lambda$, forcing the values of the entries to shrink.  As the entries are non-negative, at the end of training, the selected entries are all shrunk to zero due to the shrinking rate $\lambda$ of sufficient magnitude.  When the entries are shrunk to zero, the gradient backpropagating to earlier
layers of the saliency module will be zero, and the shrinking will stop. 
Different to abruptly pruning channels with truncation, progressive shrinking makes the network gradually adapt to the unimportant channels' degeneration and stabilizes the forward and backward propagation. Hence, the network achieves higher performance as demonstrated in Sec.~\ref{sec: Ablation Study}.

\subsection{Running Shrinking Policy}
\label{Sec: Running Shrinking Policy}
In existing dynamic channel pruning methods~\cite{dong2017more,gao2018dynamic,hua2018channel,tang2021manifold,su2020dynamic}, during forward inference, the networks can dynamically select different parameters for different inputs.
Thus, the system has to spend extra time for filter indexing (to access the parameters at different locations of the computer memory), which can lead to higher latency despite a reduction in FLOPs. 
In other words, it is cache unfriendly and causes massive indexing costs. This is because \textbf{(1)} The activated channels do not have spatial locality for every single sample: The activated and deactivated channels are mixed up and thus are almost randomly distributed on the memory page. \textbf{(2)} Furthermore, when running in batch, the activated channels among all samples within the batch are different and lack spatial locality, which further increases the memory access overhead. We would like to avoid such random access scenarios because when accessing a certain memory, the entire blocks are loaded and then the corresponding information is indexed. When the data is scattered in multiple locations, relatively more blocks are loaded and accessed,
which leads to extra MAC occupying memory bandwidth.

To address these issues, we seek to reduce the Memory Access Cost for filter indexing by conducting dynamic pruning in the training phase and selecting salient channels that can be shared among the samples during testing, which can reduce the cost of filter indexing and further improve latency.
To select the shared channels, we propose the Running Shrinking Policy that accords to the statistics of the running average on salience for each channel to generate a more stable sorting result for the top-K selection, i.e., $R$ in Eq.~\eqref{eq:PCSGrad}.

Firstly, we calculate the exponential moving average of $\boldsymbol{s}$: 
\begin{align}
\overline{\boldsymbol{s}}=(1-\alpha)\times \overline{\boldsymbol{s}} +\alpha\times \boldsymbol{s}, 
\end{align}
where $\overline{\boldsymbol{s}}$ denotes the running salience vector, which is calculated over the mini-batches of all training iterations, $\alpha$ denotes the decreasing weight
of the exponential moving average and we use the common setting of $\alpha=0.1$. 
Secondly, we perform sorting on $\boldsymbol{s}$ based on the values of $\overline{\boldsymbol{s}}$ in each optimization step and select the top-K channels for shrinking.

According to the above, the Running Shrinking Policy changes with the result of the sorting on the running average salience ($\overline{\boldsymbol{s}}$). 
At the beginning of training, the policy is dynamic because the statistics of the average is insufficient (the average is dominated by the first few iterations). As the training progresses, the final average is calculated on more training samples, which means the average will accord to global statistics more and the policy will become stabler at the later stage of training. In the final stage of training, the policy turns static and a fixed batch of channels will be shrunk. 
As the result, the positions which can generate zero salience will be static for different inputs. 
Hence we can remove the deactivated channels after training to avoid indexing operation and reduce memory access in testing. Note that, the Running Shrinking Policy still retains the dynamic pruning scheme during training so that the shrinking policy can automatically adapt to the network status and remains flexible to reduce the decrease of network performance.

We highlight that, as shown in Fig.~\ref{PCSConv}, for each $l$-th layer, the running salience vector ${\bar{\boldsymbol{s}}}^{(l)}$ is not directly inferred from ${\boldsymbol{s}}^{(l)}$ during testing, it is instead calculated based on $\boldsymbol{s}^{(l)}$ during training and \textit{fixed during testing} thereafter, i.e., after training, we fix the value of ${\bar{\boldsymbol{s}}}^{(l)}$.

\subsection{Accelerating Inference Process}
\label{Sec: Accelerating Inference Process}
In this section, we optimize the architecture of PCS to illustrate how it reduces computational complexity (MAdds) and improves the inference speed. 
For simplicity, we denote the convolution, normalization layer, activation function, and salience generator module as one layer. Assume the $l^{\text{th}}$ layer consists of $K^{(l)}\times K^{(l)}$ convolution $F^{(l)}$. 
Given the input feature maps $\boldsymbol{x}^{(l-1)}$, the output feature maps before pruning $\boldsymbol{x}^{(l)}$ can be computed as:
\begin{align}
  \label{convolution equation}
  F^{(l)}(\boldsymbol{x}^{(l-1)})= \sigma(\boldsymbol{w}^{(l)}*\boldsymbol{x}^{(l-1)} + \boldsymbol{b}^{(l)})=\boldsymbol{x}^{(l)},
\end{align}
where $*$ is the convolution operation,  $\sigma(\cdot)$ indicates the activation function such as ReLU, and $\boldsymbol{w}^{(l)}\in \mathbb{R}^{C^{(l)}\times C^{(l-1)}\times K^{(l)}\times K^{(l)}}$ and $\boldsymbol{b}^{(l)}\in \mathbb{R}^{C^{(l)}}$ are the weights and the bias, respectively.
$C^{(l-1)}$ and $C^{(l)}$ denote the number of  input and output channels of the $l^{\text{th}}$ layer. 
For better presentation, batch normalization is not included in Eq. \eqref{convolution equation} as its running mean and running std, scale $\gamma$ and shift $\beta$ can be coupled with the weights of the convolution in inference. 

To perform channel pruning, we multiply the salience vector $\boldsymbol{s}^{(l)} \in \mathbb{R}^{C^{(l)}}$ with the output feature maps $\boldsymbol{x}^{(l)}\in \mathbb{R}^{C^{(l)}\times H^{(l)}\times W^{(l)}}$ and remove the zero-valued feature maps. 
The pruned convolutional layer is formulated as:
\begin{equation}
  \begin{aligned}
  \label{eq.pruning_output}
  \boldsymbol{x}^{\prime(l)}&=\boldsymbol{s}^{(l)}\circ\boldsymbol{x}^{(l)}\backslash\{\boldsymbol{0}\}\\
  &=\boldsymbol{s}^{(l)}\circ\sigma(\boldsymbol{w}^{(l)}*\boldsymbol{x}^{(l-1)}+\boldsymbol{b}^{(l)})\backslash\{\boldsymbol{0}\},
\end{aligned}  
\end{equation}
where $\circ$ denotes element-wise product, $\backslash\{\boldsymbol{0}\}$ denotes removing the channels that only contain zero entries.
$\boldsymbol{x}^{\prime(l)}\in\mathbb{R}^{C^{\prime(l)}\times H^{(l)}\times W^{(l)}}$ denotes the pruned feature maps, where $C^{(l)}\geq C^{\prime(l)}$, i.e., the output of the $l^{\text{th}}$ layer is compressed and the complexity (MAdds) of the $l^{\text{th}}$ convolutional layer decreases. As mentioned in Sec.\ref{Sec: Running Shrinking Policy}, the distribution of zero-valued entries of current salience vector $\boldsymbol{s}^{(l)}$ will be identical to the running salience vector $\overline{\boldsymbol{s}}^{(l)}$ after training. Hence,
Eq. \eqref{eq.pruning_output} can be written as: \textcolor{blue}{}
\begin{equation}
  \begin{aligned}
  \label{eq.pruning_output_2}
  \boldsymbol{x}^{\prime(l)}
  =\boldsymbol{s}^{(l)}\circ \boldsymbol{m}^{(l)}\circ\sigma(\boldsymbol{w}^{(l)}*\boldsymbol{x}^{(l-1)}+\boldsymbol{b}^{(l)})\backslash\{\boldsymbol{0}\},
\end{aligned}  
\end{equation}
where $\boldsymbol{m}^{(l)}$ denotes the Boolean mask generated from the running salience vector
$\overline{\boldsymbol{s}}^{(l)}$ in testing as below: 
\begin{align}
    \boldsymbol{m}^{(l)}=
      \boldsymbol{1}_{\mathbb{R}_{\ne 0}}(\overline{\boldsymbol{s}}^{(l)})
\end{align}
where $\boldsymbol{1}(\cdot)$ denotes the indicator function and $\mathbb{R}_{\ne 0}$ denotes the set of non-zero real numbers. 
Due to the associativity of convolution, Eq. \eqref{eq.pruning_output_2} can be written as: 
\begin{equation}
    \begin{aligned}
  \boldsymbol{x}^{\prime(l)} =
  &\boldsymbol{s}^{(l)}\backslash\{\boldsymbol{0}\}\circ\\
  \sigma((&\boldsymbol{m}^{(l)}\circ\boldsymbol{w}^{(l)}\backslash\{\boldsymbol{0}\})*\boldsymbol{x}^{(l-1)}+\boldsymbol{m}^{(l)}\circ\boldsymbol{b}^{(l)}\backslash\{\boldsymbol{0}\}),
\end{aligned}
\end{equation}
where $\boldsymbol{m}^{(l)}$ and $\boldsymbol{w}^{(l)}$ are coupled, i.e., pruning is performed on the weights of the convolution. 
Hence, we can use $\boldsymbol{m}^{(l)}$ to permanently prune the weights and bias after training.

Denote the pruned weights as $\boldsymbol{w}^{\prime(l)}=\boldsymbol{m}^{(l)}\circ\boldsymbol{w}^{(l)}\backslash\{\boldsymbol{0}\}$, the pruned bias as $\boldsymbol{b}^{\prime(l)}=\boldsymbol{m}^{(l)}\circ\boldsymbol{b}^{(l)}\backslash\{\boldsymbol{0}\}$, and the pruned salience vector as $\boldsymbol{s}^\prime_{\ne 0}=\boldsymbol{s}^{(l)}\backslash\{\boldsymbol{0}\}$. Then the pruned convolution $F'^{(l)}$ can be represented as 
\begin{align}
  F'^{(l)}(\boldsymbol{x}^{(l-1)})=\boldsymbol{s}^{\prime(l)}_{\ne 0}\sigma(\boldsymbol{w}^{\prime(l)}*\boldsymbol{x}^{(l-1)}+\boldsymbol{b}^{\prime(l)}),
\end{align}
where $\boldsymbol{w}^{\prime(l)}\in\mathbb{R}^{C'^{(l)}\times C^{(l-1)}\times K^{(l)}\times K^{(l)}}$. Hence, the complexity of the $l^{\text{th}}$ layer decreases. Note that each pruning reduces the complexity of both current and next layers. Finally, the weights of the convolution in the $l^{\text{th}}$ layer is $\boldsymbol{w}^{\prime(l)} \in \mathbb{R}^{C'^{(l)}\times C'^{(l-1)}\times K^{(l)}\times K^{(l)}}$, and the complexity (MAdds)~\cite{howard2017efficient} of the pruned convolution is: 
\begin{align}
C'^{(l)}\times C'^{(l-1)}\times K^{(l)}\times K^{(l)}\times H^{(l)}\times W^{(l)},
\label{eq:eq13}
\end{align}
while the complexity of the original convolution is:
\begin{align}
C^{(l)}\times C^{(l-1)}\times K^{(l)}\times K^{(l)}\times H^{(l)}\times W^{(l)},
\end{align}
where $C^{\prime(l)}\le C^{(l)}$ and $C^{\prime(l-1)}\le C^{(l-1)}$. 
After training, we prune the channels based on the values of the corresponding running
salience vector entries, i.e., the convolutional filters corresponding to the zero-valued entries are removed permanently. This results in a compact model. 
\begin{figure}[tp]
  \center
  \includegraphics[height=3cm]{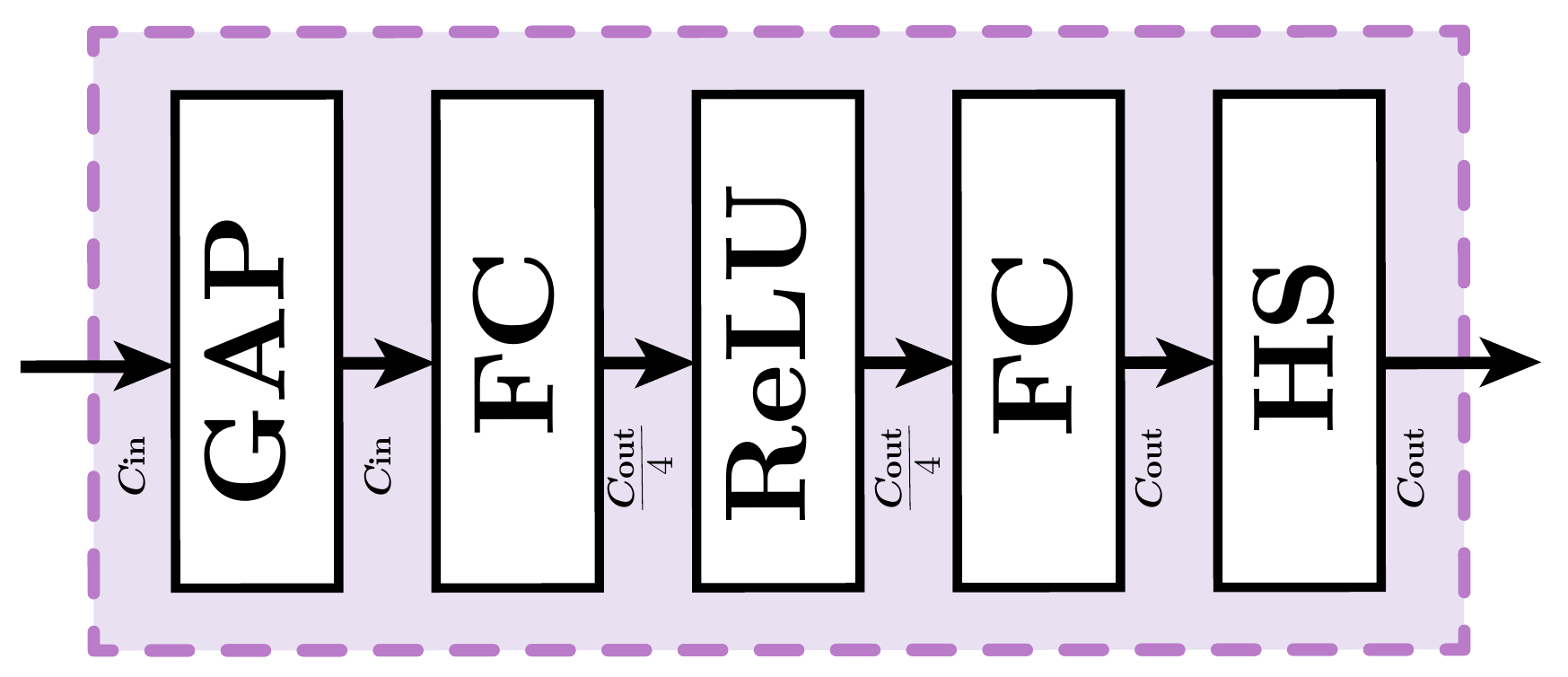} 
  \caption{
  The architecture of the Salience Generator. 
  GAP denotes global average pooling. FC denotes the fully connected layer. HS denotes hard sigmoid. }
  \label{Fig.SE}
 \vspace{-3mm}
\end{figure}

\section{Experiments}
\label{Sec: Experiments}

\begin{table*}[t]
  \begin{center}
     \caption{Comparison of SOTA channel pruning models based on ResNet-18, ResNet-34 and VGG16 on ImageNet 2012 validation dataset. MAdds and MAC denotes the number of Multiply-Add operations and memory access cost respectively. Params refers to the number of parameters in millions in the model. We also report latency measured on GPU(1080Ti) and ARM based SoC. Please refer to the Metric in Sec.~\ref{Sec: Experiments} for more details. PCS-ResNet18-A, PCS-ResNet18-B and PCS-ResNet-C denote PCS with different shrinkage $\lambda_{beta}$. * are provided by TorchVision. $\dagger$ denotes re-implementation. $\uparrow$ and $\downarrow$ denote increasing and decreasing. Lower is better for all metrics. }
     \label{Table:ResNet}
     \resizebox{0.9\textwidth}{!}{
     \begin{tabular}{lcccccccccc}
        \toprule
        \multirow{2}*{\shortstack{Model}} & Top-1 Err.& Params  & MAdds & MAC & 1080Ti & ARM & $\Delta$Top-1 Err. & $\Delta$MAdds &$\Delta$MAC\\ 
        &(\%)&(M)&(G)&(M)&(ms)&(ms)&(\%)&(G)&(M)\\
        \hline
        \textit{ResNet-18}~\cite{he2016deep} &30.2*&12 &1.8&14.5&18.9&43.2&-&-&-\\
        \textit{ResNet-18}~\cite{he2016deep} &29.6$\dagger$&12&1.8&14.5&18.9&43.2&-&-&-\\
        \textit{MIL}~\cite{dong2017more} &33.7&-&1.2&-&-&-&4.1$\uparrow$&0.6$\downarrow$&-\\
        \textit{CGNet}~\cite{hua2018channel} &31.2&12&1.0&-&-&-&1.6$\uparrow$&0.8$\downarrow$&-\\
        \textit{ManiDP-A}~\cite{tang2021manifold} & 31.1 &12& 0.9 & 15.1 &19.1 & 40.5 & 1.5$\uparrow$&  0.9$\downarrow$ & -\\
        \textit{ManiDP-B}~\cite{tang2021manifold} & 31.7 &12& 0.8 &15.1 &19.0&39.8& 2.1$\uparrow$& 1.0$\downarrow$ & -\\
        \textit{FBS}~\cite{gao2018dynamic} &31.8  & 12 &0.9&15.1&19.5$\dagger$&40.2$\dagger$&2.2$\uparrow$&0.9$\downarrow$&0.6$\uparrow$\\    
        \textit{DGC}~\cite{su2020dynamic} & 31.2 & 12 & 0.9 & 15.1 & 19.3 & 92.5 &1.6$\uparrow$&0.9$\downarrow$ & 0.6$\uparrow$\\
        \textit{\textbf{PCS-ResNet18-C (Ours)}}&\textbf{30.1}&\textbf{4}&\textbf{0.9}&\textbf{6.5}&\textbf{12.2}&\textbf{26.0}&\textbf{0.5}$\uparrow$&\textbf{0.9}$\downarrow$&\textbf{8.0}$\downarrow$\\
        \textit{\textbf{PCS-ResNet18-B (Ours)} }&\textbf{29.8}&\textbf{4}&\textbf{1.0}&\textbf{6.9}&\textbf{13.8}&\textbf{26.6}&\textbf{0.2}$\uparrow$&\textbf{0.8}$\downarrow$&\textbf{7.6}$\downarrow$\\
        \textit{\textbf{PCS-ResNet18-A (Ours)}}&\textbf{29.6}&\textbf{5}&\textbf{1.1}&\textbf{7.6}&\textbf{14.6}&\textbf{27.8}&\textbf{0}&\textbf{0.7}$\downarrow$&\textbf{6.9}$\downarrow$\\
        \hline
        \textit{ResNet-34}~\cite{he2016deep}&26.7*&22&3.6&26.9&44.0&71.0&-&-&-\\
    \textit{MIL}~\cite{dong2017more}&27.0&-&2.7&-&-&-&0.3$\uparrow$&0.9$\downarrow$&-\\
    \textit{CGNet}~\cite{hua2018channel}&28.7&22&1.8&-&-&-&2.0$\uparrow$&1.8$\downarrow$&-\\
    \textit{FBS}~\cite{gao2018dynamic}&28.3&23&1.8&28.2&29.8&77.9&1.6$\uparrow$&1.8$\downarrow$&1.3$\uparrow$\\
    \textit{ManiDP}~\cite{tang2021manifold}&27.3&23&1.7&28.2&31.1&-&0.6$\uparrow$&1.9$\downarrow$&1.3$\uparrow$\\
        \textit{\textbf{PCS-ResNet34-B (Ours)}}&\textbf{26.8}&\textbf{8}&\textbf{1.6}&\textbf{11.0}&\textbf{21.9}&\textbf{46.2}&\textbf{0.1}$\uparrow$&\textbf{2.0}$\downarrow$&\textbf{5.9}$\downarrow$\\
        \hline
        \textit{VGG16}~\cite{simonyan2014very}&28.4*&138&15.5&155&83.4&190.4&-&-&-\\
        \textit{FBS}~\cite{gao2018dynamic}&29.5&139&3.0&156&90.7&203.9&1.1$\uparrow$&12.5$\downarrow$&1.0$\uparrow$\\
        \textit{\textbf{PCS-VGG16-B (Ours)}}&\textbf{28.5}&\textbf{45}&\textbf{2.8}&\textbf{51.1}&\textbf{30.9}&\textbf{57.0}&\textbf{0.1}$\uparrow$&\textbf{12.7}$\downarrow$&\textbf{103.9}$\downarrow$\\
        \bottomrule
     \end{tabular}
     }
  \end{center}
\end{table*}

\textbf{Experimental Settings.} To evaluate PCS, we perform experiments on ImageNet dataset with VGG~\cite{simonyan2014very} and ResNet-18/34~\cite{he2016deep} on NVIDIA A100 GPUs. We further experiment on CIFAR10~\cite{krizhevsky2009learning} using ResNet-20 and ResNet-34 following the existing works~\cite{he2018soft, he2019filter, tang2021manifold}. The training settings of different networks follow their original papers. We use 60 shrinking epochs for PCS and adopt the same step learning rate policy as VGG and ResNet. We report 3 different shrinking rate policy \textbf{PCS-A}, \textbf{PCS-B} and \textbf{PCS-C}, where we set $\lambda_{base}$ in  $\lambda(T_{cur})=\lambda_{base}(\frac{T_{cur}}{T_{max}})^2$ to $4\times 10^{-6}$, $6\times 10^{-6}$, and $8\times 10^{-6}$ respectively. $T_{cur}$ and $T_{max}$ denote the current shrinking epoch and the maximum shrinking epoch. 

\textbf{Salience Generator.} 
As shown in Figure~\ref{Fig.SE}, the architecture of our salience generator is as follows: GAP$\longrightarrow$ FC$\longrightarrow$ReLU$\longrightarrow$FC$\longrightarrow$HS, where GAP, FC, and HS denote global average pooling, fully connected layer, and hard sigmoid, respectively. The salience generator takes the input feature maps of the convolutional layer as input and generates the channel-based salience vector $s$. The channel width of each fully connected layer is shown in Fig.~\ref{Fig.SE}. 
Following MobileNet V3 \cite{howard2019searching} and ShuffleNet V2 \cite{ma2018shufflenet}, the hard sigmoid is used as the activation to normalize the output vector. 

\textbf{Metrics.} 
To comprehensively evaluate the performance of our proposed method, we report not only the accuracy and computation cost but also the memory access cost (MAC) and the actual latency. Following common practices, we adopt the standard single-center crop to measure Top-1 error and use the number of MAdds as the metric of computational complexity. We calculate MAC according to~\cite{ma2018shufflenet}:
\begin{equation}
\label{eq:mac_eq1}
\begin{aligned}
    \text{MAC}=\underbrace{C_{\text{in}}\times H_{\text{in}}\times W_{\text{in}}}_{\text{input feature maps}} + \underbrace{
    C_{\text{in}}\times C_{\text{out}}\times K\times K}_{\text{convolutional kernel}}\\ + \underbrace{C_{\text{out}}\times H_{\text{out}}\times W_{\text{out}}}_{\text{output feature maps}},
\end{aligned}
\end{equation}
which consists of the MAC of the input feature maps, the MAC of the  convolutional kernel, and the MAC of the output feature maps, where $C_{\text{in}}, C_{\text{out}}$ and $K$ denote the number of input channels, the number of output channels, and the kernel size of the convolution, and $H_{\text{in}}, W_{\text{in}}, H_{\text{out}}$ and $W_{\text{out}}$ denote the height and the width of the input and output feature maps, respectively. The MAC can be considered as the footprint of the memory for the feature maps and convolutional kernels. We do not calculate the MAC for the input feature maps in Eq.~\ref{eq:mac_eq1} when adapting it to \emph{the whole network}, because the MAC of the input feature maps has been calculated as the output feature maps in the previous layer. Hence, the MAC of the whole network is calculated as:
\begin{align}
\label{eq:mac_eq2}
    \text{MAC}=\underbrace{
    C_{\text{in}}\times C_{\text{out}}\times K\times K}_{\text{convolutional kernel}} + \underbrace{C_{\text{out}}\times H_{\text{out}}\times W_{\text{out}}}_{\text{output feature maps}}.
\end{align}

We use Eq.~\ref{eq:mac_eq2} to calculate the MAC for all methods on all datasets. 
Note that, the complete convolutional kernels of other dynamic neural architectures are required to be indexed online, and hence they still occupy the memory in testing and the MAC cannot be compressed. Instead, as analysed in Section~\ref{Sec: Accelerating Inference Process} our method can offload the pruned channels for the convolutional kernels after training based on our Running Shrinking Policy, and therefore our PCS reduces MAC significantly in testing.

For latency measurements, we report the measured elapsed time on GPU (NVIDIA GTX 1080Ti) with batch size 32 and ARM CPU (Apple M1 APL1102) with batch size 1 to evaluate the benefits brought by the MAC reduction. Note that the actual latency is affected by both the computation cost (MAdds) and memory access cost (MAC). Models with lower MAdds could potentially have a larger latency due to extensive irregular access.

\begin{figure*}[t]
  \center
  \includegraphics[height=3.1cm]{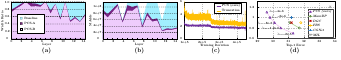}
  \vspace{-5mm}
  \caption{ (a) and (b) show the average width ratios and MAdds of PCS-A and PCS-B on ImageNet 2012 validation dataset in each layer. Baseline denotes the original ResNet-18. Layer denotes the layer index. Width Ratio denotes the channel width ratio of the pruned one to the original one, i.e.,  $\frac{C'^{(l)}}{C^{(l)}}$. (c) shows the loss of PCS and truncation during training. Our PCS achieves lower loss and is more stable than the Truncation based method. 
  (d) shows the complexity and top-1 error tradeoff with different base shrinking rates $\lambda_{base}$, which are used to construct the shrink rate policy following  $\lambda(T_{cur})~=~\lambda_{base}(\frac{T_{cur}}{T_{max}})^2$. The complexity moves inverse-proportionally to the shrinking rates. 
  It shows our PCS achieves better complexity and performance than SOTA models. }
  \label{Width Ratio}  
  \vspace{-5mm}
\end{figure*}

\subsection{ImageNet Results}
We present the experiment results on ImageNet as well as the comparisons with SOTA in Table \ref{Table:ResNet}. All PCS-A, PCS-B, and PCS-C reduce the computational complexity (MAdds) and remain a lower Top-1 error than other methods. Under similar MAdds, our PCS-ResNet18-B significantly reduces the network performance drop and achieves higher accuracy than other methods. It should be noted that the Top-1 error of our PCS is nearly equivalent to the ResNet baseline while reducing the MAdds by half. Compared to the recent SOTA methods, DGC and ManiDP-A, our PCS-ResNet18-C reduces the reduction of performance by 1.1$\%$ and 1.0$\%$ under the same MAdds, respectively. 

Moreover, our PCS decreases MAC by half while other methods slightly increase the MAC, which often bottlenecks the inference speed~\cite{ma2018shufflenet}. As the result, we can observe that our PCS reduces the latency significantly on both GPU and ARM. Due to Running Shrinking Policy, the pruning scheme is identical for all input samples. Hence, we can prune the deactivated channels before deployment to avoid extra indexing operations during inference. 
Note that ManiDP and FBS are slower than the baseline (ResNet-18) on 1080Ti because input-related pruning scheme results in multiple indexing operations for every sample in a mini-batch and introduces latency. 

Besides, we further adopt PCS-B with ResNet-34 and VGG-16. As shown in Table \ref{Table:ResNet}, both of them outperform the existing methods on complexity-performance tradeoff. Compared with the SOTA models, our PCS-ResNet34-B further reduces the MAdds with no degradation of network performance. Based on VGG-16, our PCS-VGG16-B is only one-fifth computational complexity of the baseline under the similar network performance and increases accuracy by 1\% with a lower complexity compared to FBS.

After training, the channel width of PCS-ResNet18 is fixed. Figure \ref{Width Ratio} (a) visualizes the width ratio of PCS to Baseline ResNet-18 on ImageNet 2012 validation dataset, where our PCS prunes more channels for even layers and relatively less for the odd layer. This is due to the odd layers are coupled with the residual connection. The residual connection is the key for gradient passing through the layer, and the larger channel width allows the weight behind obtaining higher bypass gradients for the update. This might be conducive to network performance and also is the result of the joint optimization.
Figure \ref{Width Ratio} (b) illustrates that the shrinkage is more significant for the wider layer where there are more redundant channels than the thinner places. 

\begin{table}[h]
  \begin{center}
      \caption{Comparison of SOTA channel pruning models based on ResNet-20 and ResNet-32 on CIFAR-10 validation dataset.
      $\uparrow$ and $\downarrow$ denote increasing and decreasing.}
     \label{Table:cifar}
     \resizebox{0.48\textwidth}{!}{
     \begin{tabular}{llcccc}
        \toprule
        \multirow{2}*{\shortstack{Model}}
        &\multirow{2}*{\shortstack{Method}} & Top-1 Err. & MAdds & $\Delta$Top-1 Err.& $\Delta$MAdds\\ 
        &&(\%)&(M)&(\%)&(M)\\
        \midrule
        \multirow{9}*{\shortstack{ResNet-20}}&\textit{Baseline}~\cite{he2016deep}&7.8&41.4&-&-\\
        &\textit{SFP}~\cite{he2018soft}&9.2&23.9&1.4$\uparrow$&17.5$\downarrow$\\
        &\textit{FPGM}~\cite{he2019filter}&9.6&19.0&1.8$\uparrow$&22.4$\downarrow$\\
        &\textit{DSA}~\cite{ning2020dsa}&8.6&20.6&0.8$\uparrow$&20.8$\downarrow$\\
        &\textit{Hinge}~\cite{li2020group}&8.2&22.6&0.4$\uparrow$&18.8$\downarrow$\\
        &\textit{DHP}~\cite{li2020dhp}&8.5&20.0&0.7$\uparrow$&21.4$\downarrow$\\
        &\textit{FBS}~\cite{gao2018dynamic}&9.0&19.2&1.2$\uparrow$&22.0$\downarrow$\\
        &\textit{ManiDP}~\cite{tang2021manifold}&\textbf{8.0}&19.0&\textbf{0.2}$\uparrow$&22.4$\downarrow$\\
        &\textit{PCS (Ours)}&\textbf{8.0}&\textbf{17.6}&\textbf{0.2}$\uparrow$&\textbf{23.8}$\downarrow$\\
        
        \midrule
        \multirow{7}*{\shortstack{ResNet-32}}&\textit{Baseline}~\cite{he2016deep}&7.3&70.1&-&-\\
        &\textit{MIL}~\cite{dong2017more}&9.3&48.2&2.0$\uparrow$&21.9$\downarrow$\\
        &\textit{SFP}~\cite{he2018soft}&9.2&41.0&1.4$\uparrow$&29.1$\downarrow$\\
        &\textit{FPGM}~\cite{he2019filter}&8.1&32.8&0.8$\uparrow$&37.3$\downarrow$\\
        &\textit{FBS}~\cite{gao2018dynamic}&8.0&31.1&0.7$\uparrow$&39.0$\downarrow$\\
        &\textit{ManiDP}~\cite{tang2021manifold}&7.9&25.8&0.6$\uparrow$&44.3$\downarrow$\\
        &\textit{PCS (Ours)}&\textbf{7.6}&\textbf{24.9}&\textbf{0.3}$\uparrow$&\textbf{45.2}$\downarrow$\\
        \bottomrule
     \end{tabular}
     }
  \end{center}
  \vspace{-5mm}
\end{table}

\begin{table}[t]
\setlength{\tabcolsep}{0.6mm}
  \begin{center}
     \caption{Comparison of using Progressive Channel-Shrinking and using Truncation on PCS-ResNet18-B. Baseline denotes ResNet-18. $\uparrow$ and $\downarrow$ denote increasing and decreasing, respectively.
     }
     \label{Table:Truncation}
     \resizebox{0.47\textwidth}{!}{
     \begin{tabular}{lcccccccc}
        \toprule
        \multirow{2}*{\shortstack{Model}} & Top-1 Err. & MAdds  & MAC & 1080Ti & ARM  &$\Delta$Top-1 Err. & $\Delta$MAdds &$\Delta$MAC\\ 
        &(\%)&(G)& (M) & (ms) & (ms) &(\%))&(G) & (M)\\
        \hline
        \textit{Baseline}~\cite{he2016deep} &29.6&1.8&14.5&18.9&43.2&-&-&-\\
        
        \specialrule{0em}{1pt}{1pt}
        \hline
            \textit{Truncation}&35.9&1.0& 15.0 & 19.4 & 40.1 &6.3$\uparrow$&0.8$\downarrow$&0.5$\uparrow$\\
            \textit{PCS}&\textbf{29.8}&\textbf{1.0}&\textbf{6.9}&\textbf{13.8}&\textbf{26.6}&\textbf{0.2}$\uparrow$&\textbf{0.8}$\downarrow$&\textbf{7.5}$\downarrow$\\
        \bottomrule
     \end{tabular}
     }
  \end{center}
  \vspace{-5mm}
\end{table}

\subsection{CIFAR10 Results}
We adopt the proposed PCS model with ResNet-20/32 and evaluate on CIFAR10 dataset.
As shown in Table \ref{Table:cifar}, the proposed PCS model reduces the computational complexity (MAdds), and its performance is also on par with the baseline models, if not better. 
Under similar Top-1 errors, our PCS module requires less MAdds, showing that our proposed PCS module can achieve a better tradeoff between computation complexity and performance, compared with the SOTA channel pruning methods.

\subsection{Ablation Study}
\label{sec: Ablation Study}
To validate the introduction of our Progressive Shrinking approach as well as the Running Shrinking Policy, we conduct extensive ablation studies. Furthermore, we investigate the trade-off between complexity and performance by ablating the shrinking rate $\lambda$. We also conduct experiments on different values of the exponential moving average factor $\alpha$ when aggregating the running salience vector $\overline{\boldsymbol{s}}^{(l)}$.

\subsubsection{Progressive Shrinking v.s. Truncation.}
\label{Table: Progressive Shrinking and Truncation}
Our method dynamically selects and progressively shrinks the low-salience channels to zero during training, hence the network can gradually adapt to the degeneration of those pruned channels. 
On the contrary, truncating the low-salience channels is so abrupt that the network performance is harmed during training. This is because the low-salience channels still contribute information to a certain extent and suddenly truncating them makes the forward propagation less stable. To evaluate the effect of truncation on the network performance, we conduct an ablation experiment with Progressive Channel-Shrinking (PCS) and truncation-based model on PCS-ResNet18-B, where we truncate $30\%$ of channels to match the computational complexity. As shown in Table \ref{Table:Truncation}, the Top-1 error of truncation increases significantly. It demonstrates that pruning the low-salience channels by truncating brings negative effects to the training and drops the network performance. Further, Fig.~\ref{Width Ratio} (c) shows the training progress of both progressive shrinking and truncation, where we observe more fluctuations during training the truncation based methods, indicating the less stable training process brought by truncation.

\begin{table}[t]
\setlength{\tabcolsep}{0.8mm}
  \begin{center}
     \caption{Comparison of using Running Shrinking Policy and using input-dependent shrinking policy on PCS-ResNet18-B. Baseline denotes ResNet-18. $\uparrow$ and $\downarrow$ denote increasing and decreasing, respectively. 
     }
     \label{Table:static&dynamic}
     \resizebox{0.47\textwidth}{!}{
     \begin{tabular}{lcccccccc}
        \toprule
        \multirow{2}*{\shortstack{Model}} & Top-1 Err. & MAdds  & MAC & 1080Ti & ARM  &$\Delta$Top-1 Err. & $\Delta$MAdds &$\Delta$MAC\\ 
        &(\%)&(G)& (M) & (ms) & (ms) &(\%)&(G) & (M)\\
        \hline
        \textit{Baseline}~\cite{he2016deep} &29.6&1.8&14.5&18.9&43.2&-&-&-\\
        \specialrule{0em}{1pt}{1pt}
        \hline
        \textit{Input-dependent Shrinking}&\textbf{29.8}& \textbf{0.9}&15.1&19.1&37.5&\textbf{0.2}$\uparrow$ & \textbf{0.9}$\downarrow$ & 0.6$\uparrow$\\
        \textit{Running Shrinking}&\textbf{29.8}&1.0&\textbf{6.9}&\textbf{13.8}&\textbf{26.6}&\textbf{0.2}$\uparrow$ &0.8$\downarrow$&\textbf{7.6}$\downarrow$\\
        \bottomrule
     \end{tabular}
     }
  \end{center}
  \vspace{-5mm}
\end{table}

\subsubsection{The Performance Drop Brought by Running Shrinking Policy.}
In this work, we use the Running Shrinking Policy during training. Unlike the input-related shrinking policy, it can select and shrink the identical channels for different input samples. After training, the proposed Running Shrinking Policy will construct a static pruning scheme, which can avoid indexing operation in inference but cannot adapt to the input samples. Contrarily, the input-related shrinking leads to a dynamic pruning scheme, and it has input-adaptive computational complexity. To evaluate the practical efficiency of the Running Shrinking Policy, we conduct an ablation experiment on these two shrinking policies. Table \ref{Table:static&dynamic} shows their effect on the network performance, complexity, and inference speed. Input-related pruning scheme achieves slightly lower MAdds with the same Top-1 error. However, its latency is 40\% higher than the static pruning scheme due to more extensive memory operations. In this case, the lower complexity becomes less meaningful and does not directly relate to the inference speed in practice. 

\subsubsection{The Performance-Complexity Tradeoff.}
In this section, we investigate the relationships between the performance of the pruned models and their computation complexity using ResNet18 on ImageNet. Figure~\ref{Width Ratio} (d) plots the correlation between performance and complexity under different base shrinking rates $\lambda_{base}$, which are used to construct the shrink rate policy following $\lambda(T_{cur})=\lambda_{base}(\frac{T_{cur}}{T_{max}})^2$. When raising the shrinking rate, the computation complexity (MAdds) reduces inverse-proportionally. We further plot the performance complexity tradeoff for SOTA models.  
Our proposed PCS achieves better performance-complexity tradeoff compared to the SOTA, indicating as both the lower MAdds and Top-1 Error.

\subsubsection{Efficient Network Results}

\begin{table}[t]
\setlength{\tabcolsep}{0.8mm}
  \begin{center}
     \caption{
     Acceleration comparison of the current SOTA models and our PCS on MobileNetV2. \textit{Baseline} is the MobileNetV2. Theoretical Acl. and Realistic Acl. denote the acceleration on MAdds and latency, respectively. 
     $\uparrow$ and $\downarrow$ denote increasing and decreasing, respectively. 
     }
    
     \label{Table:MobileNetV2}
     \scriptsize
     \resizebox{0.48\textwidth}{!}{
     \begin{tabular}{lcccccccc}
        \toprule
        \multirow{2}*{\shortstack{Model}} & Top-1 Err. & MAdds  & Latency & $\Delta$ Top-1 Err. & Theoretical Acl.& Realistic Acl.\\
        &(\%)&(M) & (ms) &(\%) &(\%)&(\%)\\
        \hline
        \textit{Baseline}\cite{he2016deep} &28.0&300&75&-&-&-\\
        \specialrule{0em}{1pt}{1pt}
        \hline
        \textit{DGC}\cite{su2020dynamic} &29.3&245&-&1.3$\uparrow$&18&-\\
        \textit{ManiDP}\cite{tang2021manifold} &30.4&147&-&2.4$\uparrow$&51&39\\
        \textit{PCS (Ours)} &\textbf{28.6}&\textbf{147}&\textbf{39}&\textbf{0.6}$\uparrow$&\textbf{51}&\textbf{48}\\
        \bottomrule
     \end{tabular}
     }
  \end{center}
  \vspace{-5mm}
\end{table}

We further evaluate the generalization of our progressive shrinking policy on the efficient network architectures. We adopt the proposed PCS model on MobileNet V2 and evaluate on ImageNet. We use the base shrinking rate as $\lambda_{base}=6e-5$ to train our PCS on MobileNet V2. 
The results are shown in  Table~\ref{Table:MobileNetV2},
 where the latency is measured on Google Pixel 1 Phone with batch size 1 (following Sandler \textit{et al.} ~\cite{sandler2018inverted}) to obtain the realistic acceleration. The results of other methods are reported by the corresponding original papers.

\subsubsection{Evaluation of the Decreasing Weight of Exponential Moving Average for our Running Shrinking Policy}

We conduct experiments on different values of the decreasing weight $\alpha$. Table~\ref{Table:alpha} shows  performance and computation complexity of the PCS model with different decreasing weights. It can be seen that our PCS model with $\alpha = 0.1$ achieves better performance-complexity tradeoff, compared to the PCS models with $\alpha = 0.5$ and $\alpha = 0.05$. We thus use $\alpha = 0.1$ as the decreasing weight of the exponential moving average in our Running Shrinking Policy.

\begin{table}[t]
\setlength{\tabcolsep}{0.8mm}
  \begin{center}
     \caption{Comparison of different exponential moving average rate $\alpha$ for our Running Shrinking Policy. 
     The latency is measured on NVIDIA GTX 1080Ti with batch size 32. $\uparrow$ and $\downarrow$ denote increasing and decreasing, respectively.  
     }
     \label{Table:alpha}
     \resizebox{0.48\textwidth}{!}{
     \begin{tabular}{lcccccccc}
        \toprule
        \multirow{2}*{\shortstack{Model}}&\multirow{2}*{\shortstack{$\alpha$}} & Top-1 Err. & MAdds  & Latency& $\Delta$Top-1 Err. & $\Delta$MAdds  & $\Delta$Latency\\
        &&(\%)&(G)&(ms)&(\%)&(G)&(ms)\\
        \hline
        \textit{Baseline} \cite{he2016deep}& - &29.6&1.8&18.9&-&-&-\\
        \hline
        \textit{PCS(Ours)}&0.5 & 30.7 &1.0& 13.8& 1.1$\uparrow$&0.8$\downarrow$&5.1$\downarrow$\\
        \textit{PCS(Ours)}&0.1 &\textbf{29.8}&1.0&13.8&\textbf{0.2}$\uparrow$&0.8$\downarrow$&5.1$\downarrow$\\
        \textit{PCS(Ours)}&0.05 &30.0&1.0&13.8&0.4$\uparrow$&0.8$\downarrow$&5.1$\downarrow$\\
        \bottomrule
     \end{tabular}
     }
  \end{center}
\vspace{-5mm}
\end{table}

\subsubsection{Visualization of Channel Shrinking}
\label{Channel Shrinking}
\begin{figure}[t]
  \center
  \includegraphics[height=3.2cm]{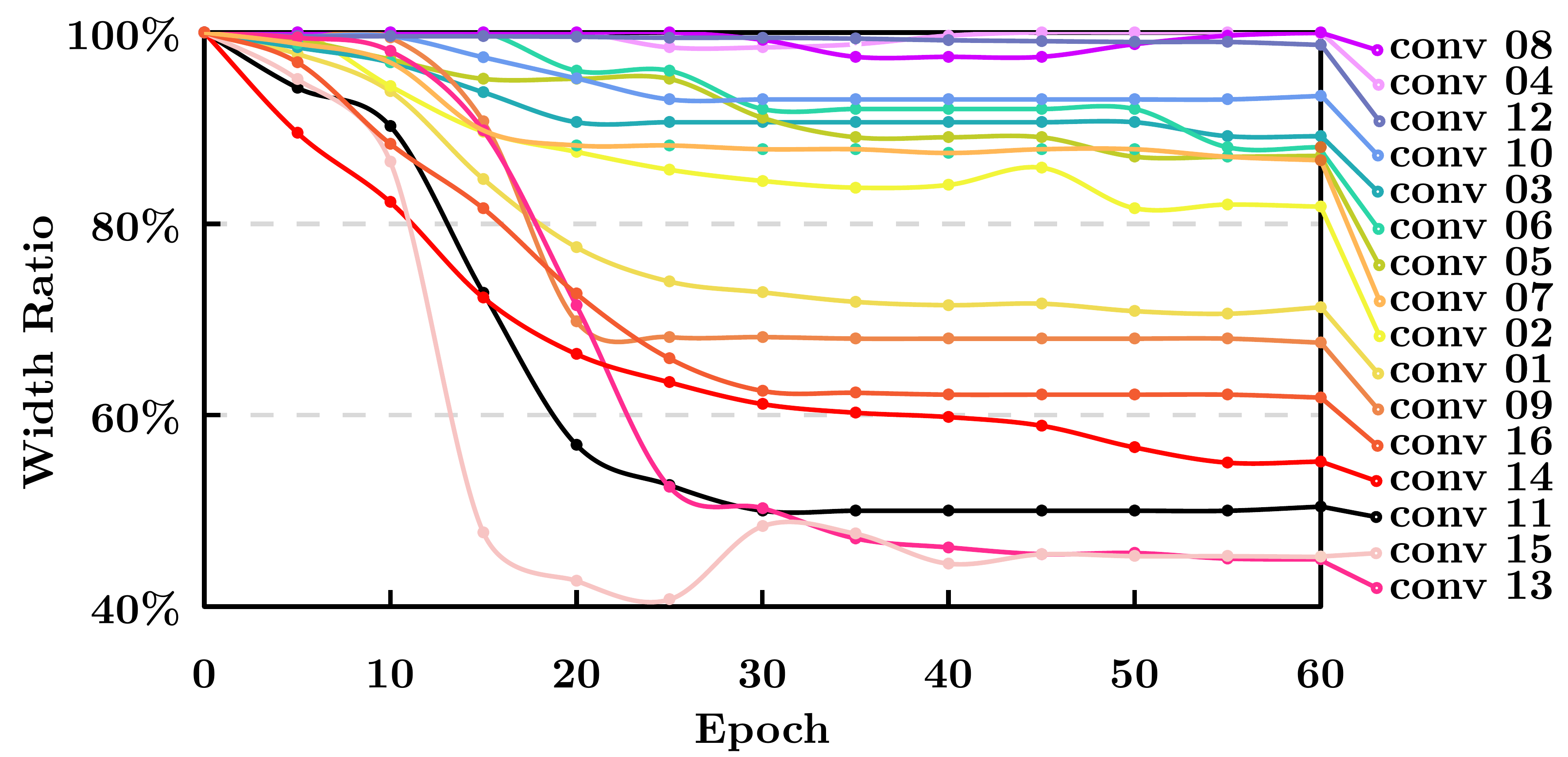}
  \caption{
  Visualization of the width ratios of different convolution layers of our PCS-Resnet18-B. Each curve illustrates the changing of width ratios for one specific convolution layer. (Best viewed in color)}
  \label{Fig: WidthChange}  
\end{figure}

We visualize the shrinking status of all the convolution layers of our PCS-ResNet18-B during the training. 
As shown in Fig. \ref{Fig: WidthChange}, the number of channels of every convolution layer is \textbf{progressively} shrunk during training. 
The shrinking speeds among various convolution layers are different in the same epoch. The shrinking speeds are also different for the same convolution layer in different epochs. This is because shrinking is a dynamic process, and the salience of channels in each layer is diverse. Hence, the task loss, which encourages the raise of the salience value of important entries to contribute more to the task-related performance, overweighs the shrinking loss, which facilitates the width shrinking among different layers. 

Note that, although the width ratios of only a few layers are decreased to below $50\%$, our PCS-ResNet18-B reduces the overall computational complexity by around $50\%$. This is because the computational complexity is quadratic with respect to the channel width ratio of each layer, e.g., when pruning all layers to $70\%$ channels, there will be only $49\%$ complexity left following Eq.~\ref{eq:eq13}. 

\subsubsection{Visualization of Entries Shrinking}
\label{Attention Entries Shrinkin}

\begin{figure}[t]
  \centering
  \includegraphics[height=3.2cm]{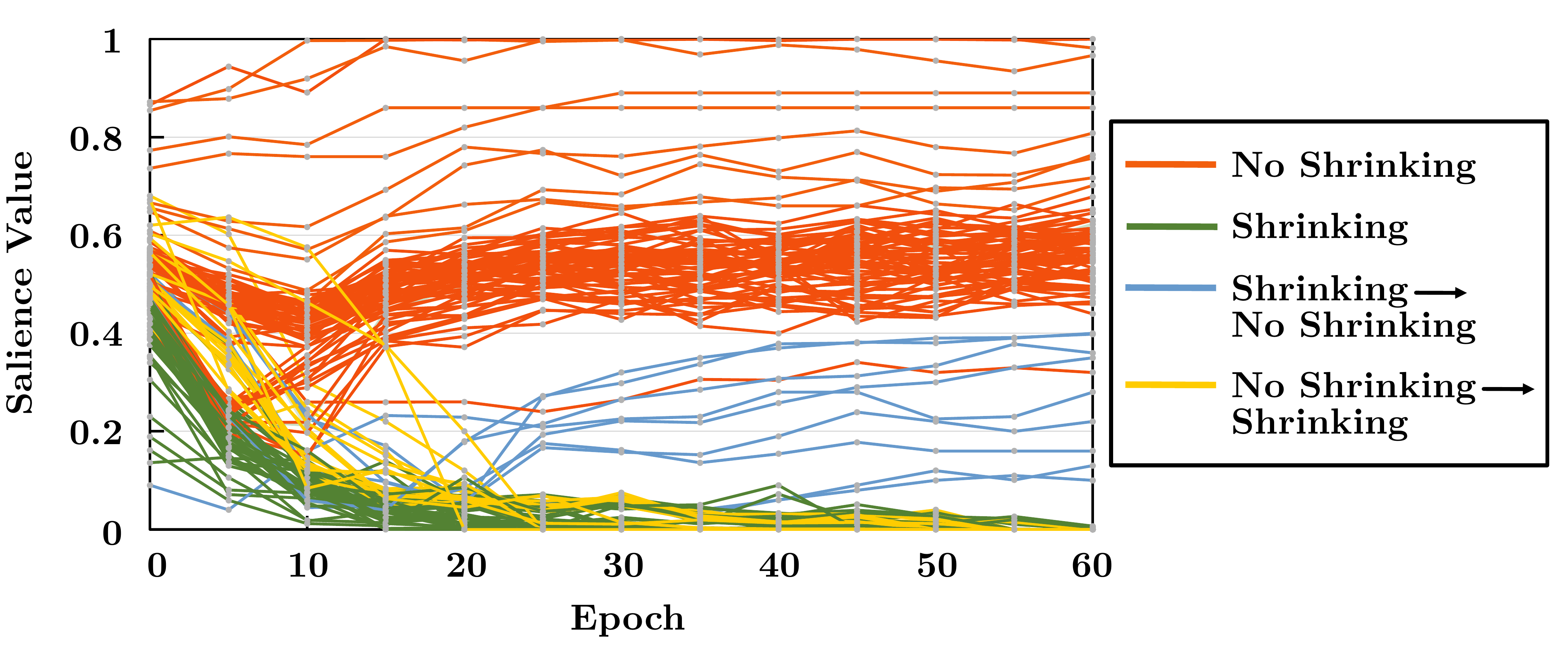}
  \caption{
  Visualization of the state of salience entries (being shrunk or not being shrunk) during training of our PCS-Resnet18-B. Each curve illustrates the value of the salience entry  during training. (Best viewed in color)}
  \label{Fig: AttentionChange}  
  \vspace{-5mm}
\end{figure}

In our proposed method, we adopt joint optimization to optimize the shrinking loss and the task loss, and use the shrinking rate to achieve the balance between compression ratio and network performance. 
At the beginning of training, the shrinking rate is relatively small, and it gradually increases alongside the training. Hence, the task-related performance is the main objective to be solved in the early stage of training. At this stage, the different entries can freely switch between the top-K lowest salience entries to be shrunk and the salient entries to be kept, which allows the model to adapt to better convergence. 
As shown in Fig. \ref{Fig: AttentionChange}, the values of salience entries are influenced during training by both the shrinking loss and the task loss, i.e., being compressed by shrinking loss to improve efficiency and meanwhile being enhanced by the task loss to improve accuracy. 
We observe that changing of salience entries can be generally summarized into 4 cases: (1) ``No Shrinking" where salience values do not shrink to a low value throughout. (2) ``Shrinking" where salience values shrink to a low value since the start.
(3) ``Shrinking $\to$ No Shrinking" where the salience value is low at the early stage of training and but in the late stage, increases to become outside of the K lowest salience values (because they contribute to good model performance and thus are increased in importance by the task loss), such that they will not be shrunk and are not pruned away. 
(4) ``No Shrinking $\to$ Shrinking" where the salience value is high at the early stage, but decreases at a later stage during training due to the effects of the shrinking loss and thus are effectively pruned away. 

Hence the salience entry fluctuates to being higher or lower than the top-$K$ lowest entries, which leads to different sorting results and makes the set of shrinking entries become dynamic. With the design of gradually increased shrinking rate, the neutral entries which are not high or low enough in salience could explore different update directions during training, allowing the model to converge to a better accuracy and cost tradeoff. Finally, at the subsequent epochs, 
the shrinking rate increases to sufficient magnitude to shrink the less salient entries to zero to actually prune the corresponding channels. 

\subsubsection{Evaluation of Different Settings of Top-$K$ for Shrinking Selection}
We conduct experiments on different values of the $K$ for top-$K$ selection in Sec.III A. Table~\ref{tab:ablation_k} shows performance and computation complexity of the PCS model with different $K$. We observe that our PCS model with $K = \frac{1}{2}C_\text{out}$ achieves a good performance-complexity tradeoff, compared to the PCS models with $K = \frac{1}{4}C_\text{out}$ and $K = \frac{3}{4}C_\text{out}$. We thus set $K = \frac{1}{2}C_\text{out}$ for the top-$K$ selection in our Progressive Shrinking.

\subsubsection{Transformer Result }
We further evaluate our PCS on transformer-like architectures. We adopt the proposed PCS model on T2T-ViT \cite{Yuan_2021_ICCV} and Swin \cite{Liu_2021_ICCV} and evaluate on ImageNet. 
As shown in Table~\ref{Table.Transformer}, our PCS can also be effective for transformer architectures.

\subsubsection{Evaluation of the Impact of Salience Generator }
We conduct an ablation experiment to evaluate the impact of the Salience Generator as shown in Table~\ref{tab:salience_generator}, where the baseline without Salience Generator denotes the original ResNet-18. Results show that using the Salience Generator brings an improvement to network performance with a slight increase in latency.

\begin{table}[h]
\footnotesize
\caption{
Performance comparison of our PCS with different $K$ values for the Progressive Shrinking.
}
  \resizebox{0.5\textwidth}{!}{
     \begin{tabular}{lcccccccc}
        \toprule
        \multirow{2}*{\shortstack{Model}} & Top-1 Err. & MAdds  & 1080Ti &$\Delta$Top-1 Err. & $\Delta$MAdds\\ 
        &(\%)&(G)& (ms) &(\%)&(G)\\
        \hline
        \textit{Baseline}&
        29.6&1.8&18.9&-&-\\
        \specialrule{0em}{1pt}{1pt}
        \hline
        \textit{$0$}&28.8& 1.8&20.3&0.8$\downarrow$&0\\
        \textit{$\frac{1}{4} C_\text{out}$}&29.2& 1.4&16.0&0.4$\downarrow$&0.4$\downarrow$\\
        \textit{$\frac{1}{2} C_\text{out}$}&29.8&1.0&13.8&0.2$\uparrow$ &0.8$\downarrow$\\
        \textit{$\frac{3}{4} C_\text{out}$}&30.3&0.5&11.4&0.7$\uparrow$ &1.3$\downarrow$\\        
        \bottomrule
     \end{tabular}
     }
     \label{tab:ablation_k}
\end{table}

\begin{table}[h]
\footnotesize
\caption{
Performance comparison of our PCS on two new transformer-based architectures.
}
  \resizebox{0.48\textwidth}{!}{
     \begin{tabular}{lccccccc}
        \toprule
        \multirow{2}*{\shortstack{Model}} & Top-1 Err. & MAdds & 2080Ti &$\Delta$Top-1 Err. & $\Delta$MAdds\\ 
        &(\%)&(G)& (ms) &(\%)&(G)\\
        \hline
        \textit{T2T-ViT}&18.5& 13.8&1.62 & - & - \\
        \textit{T2T-ViT-PCS}&18.5& \textbf{3.9}&\textbf{1.18} & 0 & \textbf{6.9}$\downarrow$\\
        \hline
        \textit{Swin}&18.7&4.5&1.32&-&-\\
         \textit{Swin-PCS}&18.7& \textbf{2.5}&\textbf{0.89} & 0 & \textbf{2.0}$\downarrow$\\        
        \bottomrule
     \end{tabular}
     }
     \label{Table.Transformer}
\end{table}

\begin{table}[h]
\footnotesize
\caption{
Performance comparison with and without the Salience Generator on baseline (ResNet-18).
}
  \resizebox{0.48\textwidth}{!}{
     \begin{tabular}{lcccccccc}
        \toprule
        \multirow{2}*{\shortstack{Model}} & Top-1 Err. & MAdds & 1080Ti & $\Delta$Top-1 Err. \\ 
        &(\%)&(G)& (ms) &(\%)\\
        \hline
        \textit{Baseline without Salience Generator. }&
        29.6&1.8&18.9&-\\
        \specialrule{0em}{1pt}{1pt}
        \textit{Baseline with Salience Generator}&28.8& 1.8&20.3 & 0.8$\downarrow$\\
        \bottomrule
     \end{tabular}
     }
     \label{tab:salience_generator}
\end{table}

\section{Conclusion}
This work introduces Progressive Channel-Shrinking network that selects and shrinks the lowest salience channels according to inputs. Rather than direct truncation, it can make the pruning operation `milder' to stabilize training. We also propose Running Shrinking Policy to reduce indexing for network acceleration. Our proposed Running Shrinking Policy makes the shrinking selection identical to all the inputs so that it can generate a static pruning scheme in testing to avoid extra indexing operations. Besides, the Running Shrinking Policy is dynamic during training to adapt the network parameters and the training status. 
The experiments show that our proposed method achieves SOTA in terms of compression-performance tradeoff and surpasses current SOTA methods on theoretical improvement. 
Furthermore, our method reduces both the FLOPs and MAC, which significantly accelerates CNNs in practice.
However, there is still potential for further optimization by combining MAC and FLOPs through hybrid optimization, and we plan to pursue this in our future research efforts.

\noindent
\textbf{Acknowledgments.}
This work is supported by MOE AcRF Tier 2 (Proposal ID: T2EP20222-0035), National Research Foundation Singapore under its AI Singapore Programme (AISG-100E-2020-065), and SUTD SKI Project (SKI 2021\_02\_06).
This work is also supported by TAILOR, a project funded by EU Horizon 2020 research and innovation programme under GA No 952215.

\bibliographystyle{splncs04}
\bibliography{egbib}
\begin{IEEEbiography}[{\includegraphics[width=1in,height=1.25in,clip,keepaspectratio]{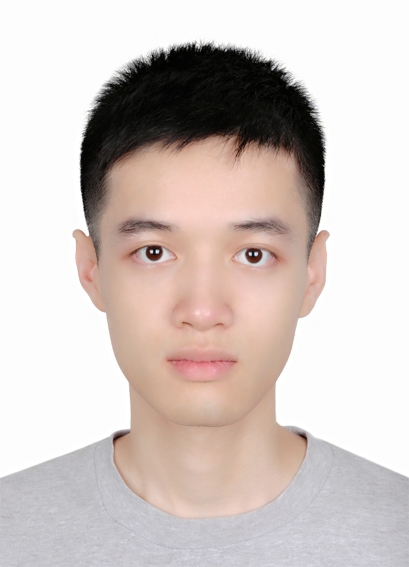}}]{Jianhong Pan} received the Bachelor degree from Shenzhen University. He currently is a research assistant in Singapore University of Technology and Design. His research interests include computer vision, object detection, efficient networks, adversarial learning, and self-supervised learning.
\end{IEEEbiography}

\begin{IEEEbiography}[{\includegraphics[width=1in,height=1.25in,clip,keepaspectratio]{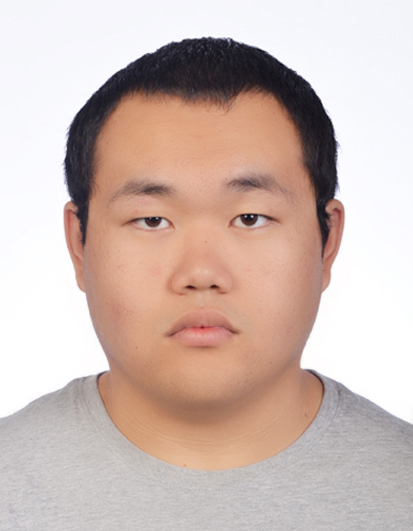}}]{Siyuan Yang} received the BEng degree from Harbin Institute of Technology and the MSc degree from Nanyang Technological University. He is currently pursuing the Ph.D. degree with the Interdisciplinary Graduate Programme, Nanyang Technological University. His research interests include computer vision, action recognition, and human pose estimation.  
\end{IEEEbiography}

\begin{IEEEbiography}[{\includegraphics[width=1in,height=1.25in,clip,keepaspectratio]{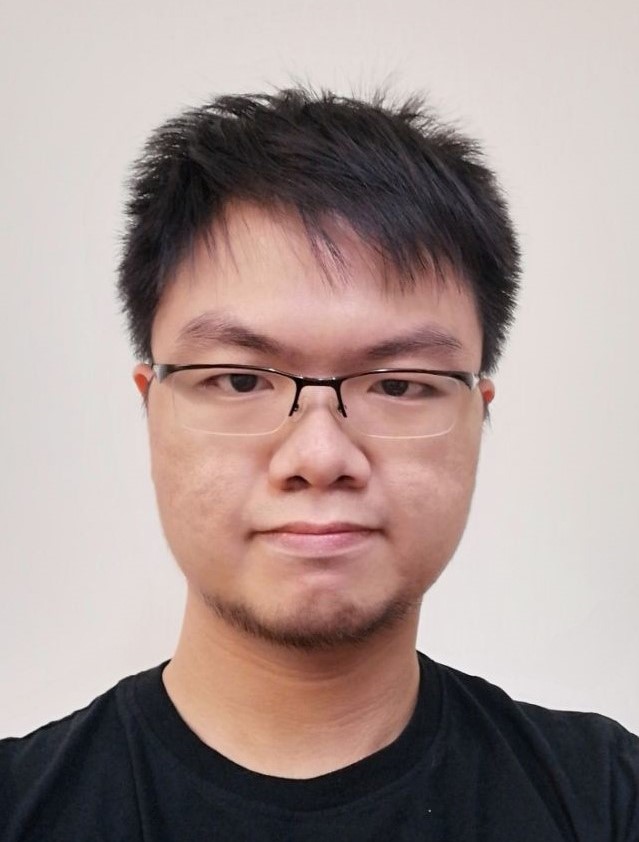}}]{Lin Geng Foo}
received the Bachelors of Engineering (Engineering Systems and Design) degree from Singapore University of Technology and Design (SUTD) in 2019. He is currently pursuing his PhD in the Information Systems Technology and Design (ISTD) pillar at SUTD. His research interests include video analysis, dynamic neural networks, and statistics.
\end{IEEEbiography}

\begin{IEEEbiography}[{\includegraphics[width=1in,height=1.25in,clip,keepaspectratio]{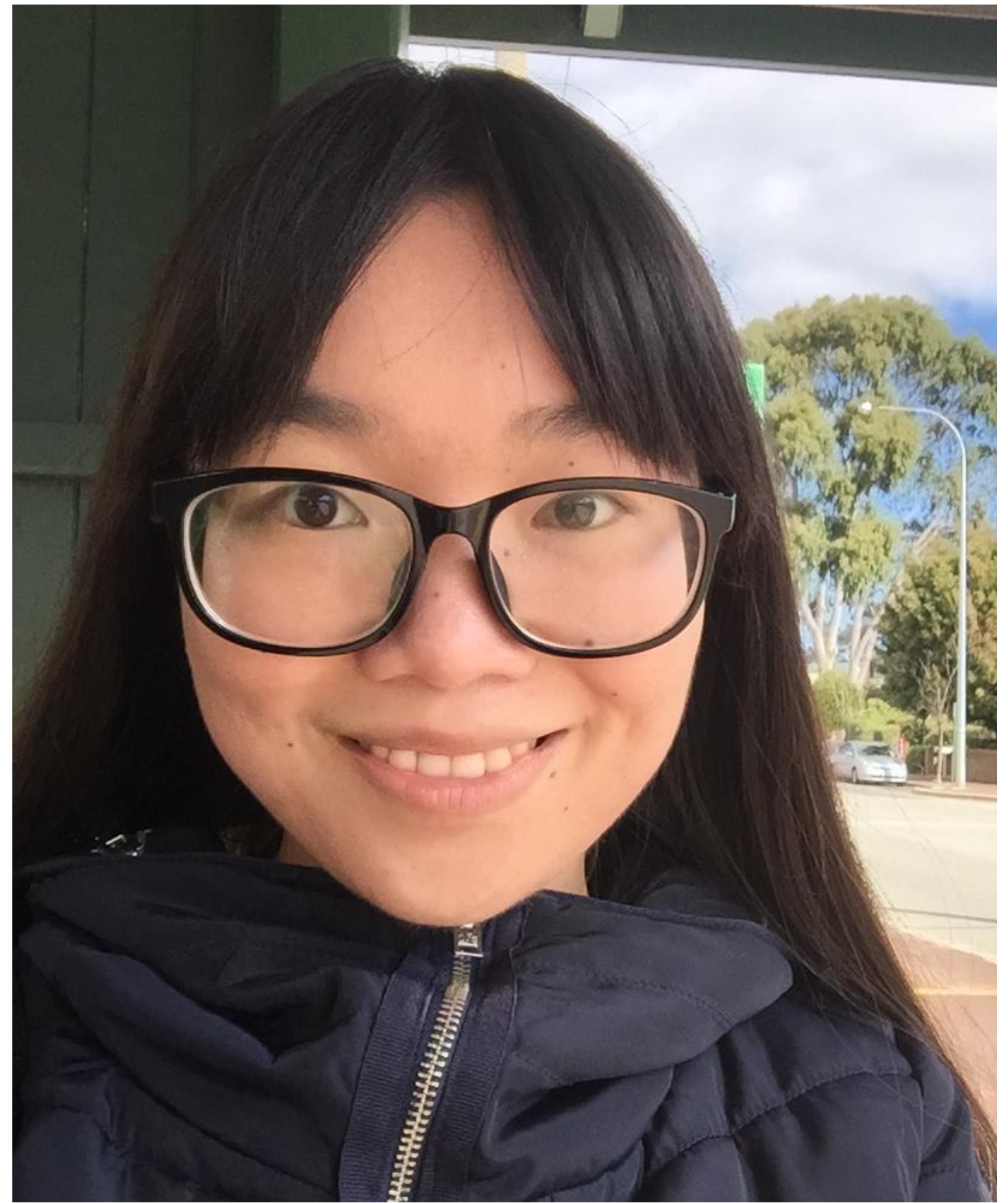}}]{Qiuhong Ke}  received her PhD degree from The University of Western Australia in 2018. She is a Lecturer (Assistant Professor) at Monash University. Before that, she was a Postdoctoral Researcher at Max Planck Institute for Informatics and a Lecturer at University of Melbourne. Her research interests include computer vision and machine learning.
\end{IEEEbiography}

\begin{IEEEbiography}[{\includegraphics[width=1in,height=1.25in,clip,keepaspectratio]{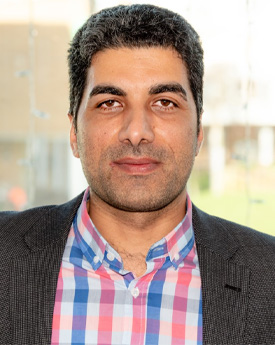}}]{Hossein Rahmani} received the B.Sc. degree in computer software engineering from the Isfahan University of Technology, Isfahan, Iran, in 2004, the M.Sc.
degree in software engineering from Shahid Beheshti
University, Tehran, Iran, in 2010, and the Ph.D.
degree from The University of Western Australia,
Perth, WA, Australia, in 2016. He is an Associate Professor (Senior Lecturer) with the School of Computing and Communications at Lancaster University in the UK. Before that, he was a Research
Fellow with the School of Computer Science and
Software Engineering, The University of Western Australia. His research interests include computer vision, action recognition, pose estimation, and deep learning.  
\end{IEEEbiography}

\begin{IEEEbiography}[{\includegraphics[width=1in,height=1.25in,clip,keepaspectratio]{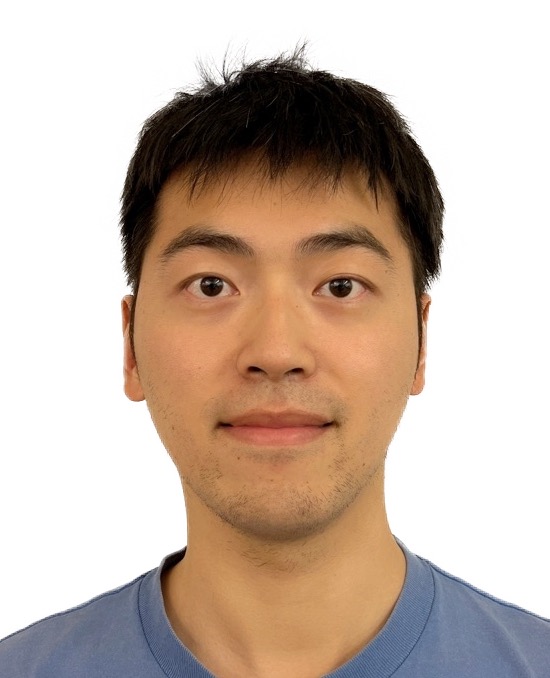}}]{Zhipeng Fan} received his Ph.D degree at Tandon School of Engineering, NYU. Previously, he was an undergraduate student at School of Precision Instruments and Opto-Electronic Engineering at Tianjin University. His research interest includes computer vision, deep learning as well as its applications.
\end{IEEEbiography}

\begin{IEEEbiography}[{\includegraphics[width=1in,height=1.25in,clip,keepaspectratio]{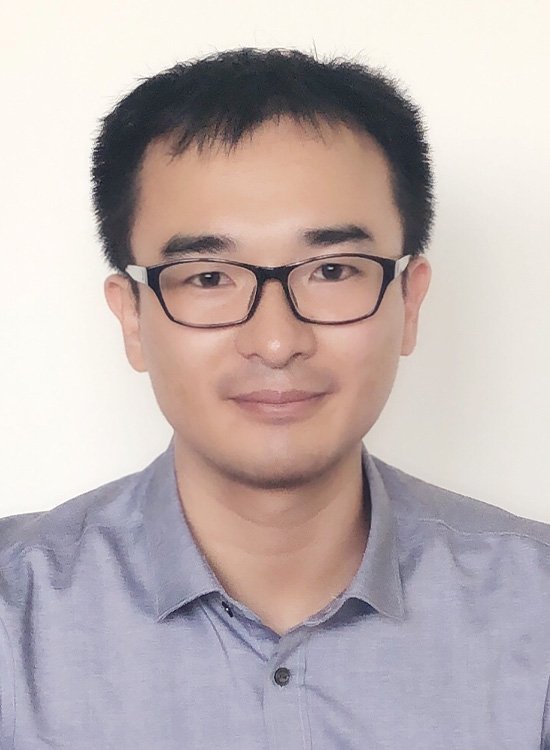}}]{Jun Liu} is an Assistant Professor with Singapore University of Technology and Design. He received the PhD degree from Nanyang Technological University, the MSC degree from Fudan University, and the BEng degree from Central South University. His research interests include computer vision and artificial intelligence. He is an Associate Editor of IEEE Transactions on Image Processing, and area chair of ICLR, ICML, NeurIPS, and WACV in 2022 and 2023.\end{IEEEbiography}

\end{document}